\documentclass[aapm,mph,preprint,graphicx]{revtex4-1}

\usepackage{graphicx}
\usepackage{multirow}
\usepackage{amsfonts,amsmath, amsbsy, amssymb}
\usepackage{epsfig}
\usepackage{hyperref}
\usepackage[displaymath,mathlines]{lineno}
\usepackage{algorithmicx,algpseudocode}

\newcommand{\Ab}{{\mathbf A}}

\newcommand{\Xb}{{\mathbf X}}
\newcommand{\Yb}{{\mathbf Y}}

\newcommand{\xb}{{\mathbf x}}
\newcommand{\yb}{{\mathbf y}}

\newcommand{\Rd}{{\mathbb R}}

\begin{document}

\title{A Deep Convolutional Neural Network using Directional Wavelets for Low-dose X-ray CT Reconstruction}

\author{Eunhee Kang$^{1,2}$, Junhong Min$^{1,2}$ and Jong Chul Ye$^{1*}$}

\affiliation{$^{1}$Bio Imaging and Signal Processing Lab. Dept. of Bio and Brain Engineering, KAIST, Daejeon, Republic of Korea\\
			 $^{2}$Contributed equally to this work}

\begin{abstract}

\textbf{Purpose:}
Due to the potential risk of inducing cancer, radiation exposure by X-ray CT devices should be reduced for routine patient scanning.
However, in low-dose X-ray CT, severe artifacts typically occur due to photon starvation, beam hardening, and other causes, all of which decrease the reliability of the diagnosis.
Thus, a high-quality reconstruction method from low-dose X-ray CT data has become a major research topic in the CT community.
Conventional model-based de-noising approaches are, however,  computationally very expensive, and image-domain de-noising approaches cannot readily remove CT-specific noise patterns.
To tackle these problems, we want to develop  a new low-dose X-ray CT algorithm based on a deep-learning approach.

\textbf{Method:}
We propose an algorithm which uses a deep convolutional neural network (CNN) which is applied to the wavelet transform coefficients of low-dose CT images.  
More specifically, using a directional wavelet transform to extract the directional component of artifacts and exploit the intra- and inter- band correlations, our deep network can effectively suppress CT-specific noise.  
In addition, our CNN is designed with a residual learning architecture for faster network training and better performance. 

\textbf{Results:}
Experimental results confirm that the proposed algorithm effectively removes complex noise patterns from CT images derived from a reduced X-ray dose.
In addition, we show that the wavelet-domain CNN is efficient when used to remove noise from low-dose CT compared to existing approaches.
Our results were rigorously evaluated by several radiologists at the Mayo Clinic and won second place at the 2016 ``Low-Dose CT Grand Challenge.''

\textbf{Conclusions:}
To the best of our knowledge, this work is the first deep-learning architecture for low-dose CT reconstruction which has been rigorously evaluated and proven to be effective.
In addition, the proposed algorithm, in contrast to existing model-based iterative reconstruction (MBIR) methods, has considerable potential to benefit from large data sets.
Therefore, we believe that the proposed algorithm opens a new direction in the area of low-dose CT research.

\end{abstract}

%\pacs{}

\maketitle

\section{Introduction}

X-ray computed tomography (CT) is a commonly used medical imaging method capable of showing fine details inside the human body, such as structures of the lung and bones.
However, CT is inherently associated with a much higher X-ray dose compared to simple film radiography.
Due to the strong evidence of radiation-related cancer \cite{brenner2007computed}, the recent research interest in CT has mainly focused on minimizing the X-ray dose to reduce the risk to patients \cite{yu2009radiation}.
One popular technique by which to do so is to reduce the number of X-ray photons emitted from the X-ray source by controlling the currents applied to the X-ray tube.
However, such a technique typically results in reduced image quality due to the low signal-to-noise ratio (SNR) measurements.
Accordingly, the success of low-dose CT is strongly determined by the de-noising technique used.

There are various image de-noising approaches.
Popular approaches include total variation minimization \cite{chambolle2004algorithm} and wavelet shrinkage approaches \cite{portilla2003image,crouse1998wavelet}.
More recent methods use non-local statistics or the degree of self-similarity of the images \cite{dabov2007image, kim2015sparse}.
However, one of the limitations of existing image-domain de-noising approaches is that they are not ideal when used with CT-specific noise patterns, as complicated streaking artifacts usually occur in low-dose CT due to photon starvation and beam hardening \cite{hsieh2003computed}.
In addition, CT data are subject to sophisticated non-linear acquisition processes, which lead to non-stationary and non-Gaussian noise processes.
To address these limitations, model-based iterative reconstruction (MBIR) was developed. 
It relies on the physical modelling of projection/backprojection operators and the statistical modelling of the noise in projection measurements \cite{beister2012iterative}.

Researchers of MBIR algorithms have developed optimization techniques to speed up  the convergence of MBIR\cite{ramani2012splitting,kim2015combining} as well as new regularization terms  that can reduce the noise and preserve the edge details\cite{xu2012low,wieczorek2015x,zhang2015statistical,cho2015regularization}.
Although the existing MBIR approaches may include a system geometry model that takes the CT scanner geometry and physical effects into account, MBIR approaches typically perform computationally expensive iterative projection/backprojection steps.
In addition, it is difficult for MBIR to use the rich information available in large-scale CT data sets, as only a few parameters can be trained in typical MBIR algorithms.

In computer vision applications, de-noising algorithms using an artificial neural network have been intensively studied and have shown impressive performance capabilities \cite{zhang2005image,jain2009natural,nasri2009image,vincent2010stacked,burger2012image,xie2012image,mao2016image,chen2015learning}.
In this de-noising framework, the parameters of a neural network are trained by supervised learning using large training data sets and the trained network is then applied to remove noise from the test data set.
Although classical neural network approaches were limited to shallow structures due to gradient vanishing/exploding or overfitting problems, the recent development of new network units, such as the rectified linear unit(ReLU), max pooling, dropout and batch normalization, mitigate the problems associated with classical methods, leaving much deeper networks with more power.
For the last few years, researchers have had great successes from deep networks in many low-level computer vision applications, such as de-noising \cite{xie2012image,mao2016image,chen2015learning} and super-resolution applications \cite{dong2014learning,kim2016accurate}.

Inspired by the success of the deep convolutional neural network,  we propose a novel low-dose CT de-noising framework designed to detect and remove  CT-specific noise patterns. 
Specifically, instead of using the publicly available image-domain CNN architecture, we propose a new CNN architecture optimized for CT de-noising.
In particular, based on the observation that a directional wavelet transform can detect the directional components of noise, we construct a deep CNN network in the wavelet domain.
More specifically, the network is trained with  wavelet coefficients from the CT images after applying the contourlet transform\cite{zhou2005nonsubsampled}.
To achieve the best performance, the proposed wavelet-domain network consists of operation units such as convolution, batch normalization\cite{ioffe2015batch, ioffe2017batch}, and rectifier linear unit (ReLU) \cite{nair2010rectified} with residual learning\cite{he2016deep} using various types of bypass connections. 

The performance of the proposed de-noising framework was rigorously evaluated using the data set of the 2016 Low-Dose CT Grand Challenge \cite{lowDose2016} and showed significant improvements compared to conventional de-noising approaches.
Extensive experimental results also confirmed the effectiveness of the proposed network architecture.

\begin{figure}
\begin{center}
\includegraphics[width=15cm]{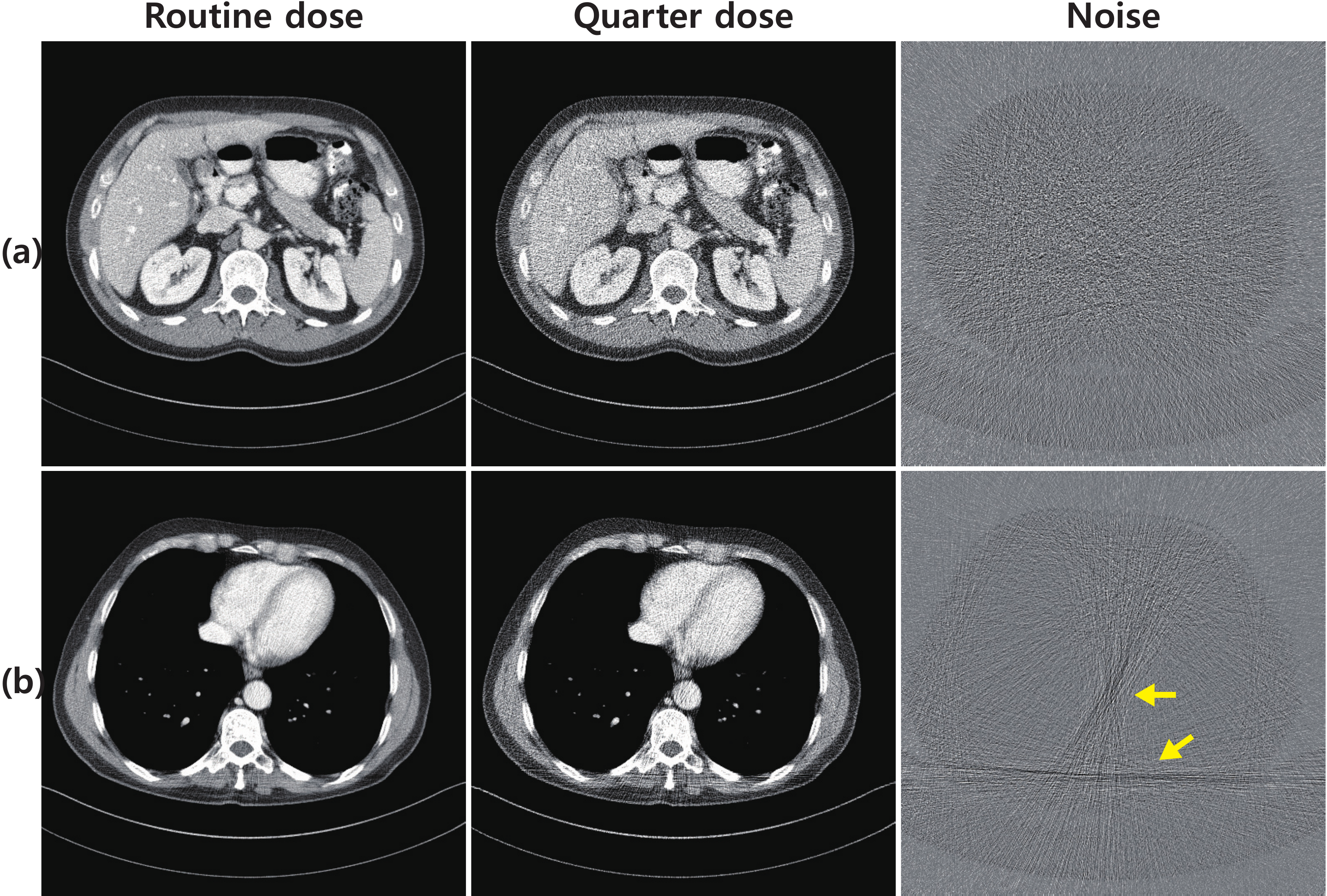}
\caption{Various noise patterns in low-dose CT images: (a) Gaussian noise, and (b) streaking artifacts}
\label{fig:low_dose_CT_image}
\end{center}
\end{figure}

\section{Background}
\label{sec:background}

\subsection{Low-dose X-ray CT Physics}

The statistics of X-ray measurements are often described by a Poisson distribution.
Specifically, a Poisson model for the intensity measurement is
\begin{equation}
	I_i \sim Poisson\left\lbrace b_i e^{-[\Ab\xb]_i} + r_i\right\rbrace, \quad i=1,\cdots,N_m ,
\end{equation}
where $\Ab$ is a system matrix (projection operator), $\xb=(x_1,\cdots,x_{N_v}) \in \Rd^{N_v}_{+}$ is a vector for the representation of  attenuation coefficients with units of inverse length, $N_m$ is the number of measurements, $N_v$ is the number of image voxels, $b_i$ denotes the X-ray source intensity of the $i$th ray, and $r_i$ denotes the background contributions of scatter and electrical noise.
After taking the log, the sinogram data is often approximated as a weighted Gaussian:
\begin{equation}
\label{equ:gaussian_model}
	y_i \sim N\left( [\Ab\xb]_i, \dfrac{\bar{I}_i}{(\bar{I}_i - r_i)^2} \right),
\end{equation}
where $\bar I_i = E[I_i]$.

In practice, polychromatic X-ray can produce various artifacts such as beam-hardening.
Given that the lower energy X-ray photons are seldom measured by detectors, the linearity of the projection data and attenuation coefficients is no longer valid.
In addition to these beam-hardening related artifacts, there are photon-starvation artifacts because bones have higher attenuation and thus absorb a considerable amount of  X-ray photons.
Specifically, ribs and backbone are located on opposite sides of the body;
therefore, an X-ray beam should pass through more than two bones depending on the direction of the X-ray path such that we lose information about the tissues between these bones.
This results in streaking artifacts in these directions (see Fig.~\ref{fig:low_dose_CT_image}).

\subsection{Conventional algorithms for low-dose X-ray CT}

\subsubsection{Image domain de-noising}

One of the simplest approaches to low-dose X-ray CT is image-domain de-noising.
Among various approaches, wavelet shrinkage approaches which decompose an image into low- and high-frequency components with thresholding for the high-frequency coefficients have been widely used \cite{portilla2003image}.
Advanced algorithms in this field exploit the intra- and inter- correlations of the wavelet coefficients of image details by statistical modeling \cite{crouse1998wavelet}.
Wavelet shrinkage approaches indeed correspond to the application of a sparsity penalty for wavelet transform coefficients.
Accordingly, sparsity-driven de-noising algorithms have been extensively studied, with the approach known as total variation (TV) widely used \cite{chambolle2004algorithm}.
Unfortunately, this approach often produces cartoon-like artifacts.

To solve this problem, some studies approximate noisy patches using a sparse linear combination of the elements of a learned dictionary\cite{elad2006image}.
Newer approaches use the non-local statistics of images based on the observation that different local patches in the same image are often similar in appearance.
For example, block matching with a 3D collaborative filtering algorithm (BM3D) \cite{dabov2007image} uses the self-similarity of small patches and applies group-based filtering to similar patches to suppress noise. 
However, these methods have not been designed directly  for X-ray CT artifacts.
Therefore, it is often not possible to detect and remove CT-specific noise features.

\subsubsection{MBIR for low-dose X-ray CT}

To address these issues, model-based iterative reconstruction (MBIR) approaches have been studied extensively. 
Specifically, using the model  in  Eq.~\eqref{equ:gaussian_model}, this problem is formulated as the following minimization problem:
\begin{equation}
	 \min_{\xb} \left\lbrace \dfrac{1}{2} \|\yb - \Ab\xb\|^2_w + \lambda R(\xb) \right\rbrace,
\end{equation}
where $\|\yb - \Ab\xb\|^2_w$ is a data fidelity term weighted by $\bar I_i/(\bar I_i -r_i)$, $R(\xb)$ is a regularization term (penalty term) that imposes additional requirements such as smoothness or sparsity, and $\lambda$ is a regularization parameter.

A popular regularization term is again total variation (TV) \cite{chambolle2004algorithm} regularization, based on the assumption that an image  under  gradient operation is sparse.
Many researchers have developed different types of dictionary or non-local means methods.
For example, a dictionary that represents image features such as edges is constructed or updated during the reconstruction process\cite{xu2012low}.
In addition, existing ridgelets, shearlets, and curvelets can compose the dictionary \cite{wieczorek2015x}.
Other methods based on non-local means regularization have also been developed \cite{zhang2015statistical,cho2015regularization}.
Although there have been extensive studies to speed up  convergence\cite{ramani2012splitting,kim2015combining}, these iterative reconstruction methods are on the other hand associated with high computational complexity of the projection/backprojection operators and iterative optimization steps for non-differentiable sparsity promoting penalties.

\subsection{Convolutional neural networks}

Dramatic improvements in parallel computing techniques allow the processing of large amounts of data for deep neural networks. 
The recent breakthroughs in deep neural networks originated from deep convolutional neural networks (CNNs) such as AlexNet \cite{krizhevsky2012imagenet}.  
The convolutional neural network, inspired by the neural network of the visual cortex in animals, is a special case of an artificial neural network. 
Similar to typical neural networks, it consists of successive linear and non-linear functions, but the linear parts are specifically expressed by convolution operations. 
In particular, the local property of the convolution is known to be efficient with regard to understanding visual data, and the induced nonlinearity allows for far more complex data representations.  
The deeper the network becomes, the greater the abstraction of images.  

The simplest form of the CNN output $y$ is expressed as
\begin{equation}
	y= F(\Theta,x) =  f_n ( W_n f_{n-1}( \cdots(f_2(W_2 f_1(W_1 x + b_1 )+b_2)\cdots) + b_{n}), %\quad k=1,\cdots, K, 
\end{equation}
where $x$ is the input, $y$ is the output, $W_i$ is the convolution matrix of the $i$-th layer, $b_i$ is the bias of the $i$-th convolution layer, $f_i$ is a nonlinear function, and $\Theta$ is the set of all tunable parameters including $W_i$ and $b_i$. 
Although there are several non-linearity functions, the rectified linear unit (ReLU)\cite{nair2010rectified}, i.e., $f(x)=\max(x,0)$, is commonly used in modern deep architectures. 
The goal of the CNN framework is then to find an optimal parameter set $\Theta$ with $K$ inputs to minimize the empirical loss: 
\begin{equation}\label{eq:min}
%\Theta^* =    \underset{\Theta}{\mathrm{argmin}}
\sum_{k=1}^K L\left(y_k, F(\Theta, x_k) \right). 
\end{equation}
In this equation, $x_k$ and $y_k$ denote the $k$-th input and output, respectively.
Here, $L$ typically denotes the cross-entropy loss in classification problems or the Euclidean distance in regression problems such as image de-noising. 
Assuming that all nonlinear functions and the loss functions are differentiable, the minimization problem or network training in \eqref{eq:min} can be addressed by an error back-propagation method  \cite{cotter2011better}.  
In general, the performance of the deep network is determined by the network architecture and methods that overcome the overfitting problem.

\begin{figure}
\begin{center}
\includegraphics[width=12cm]{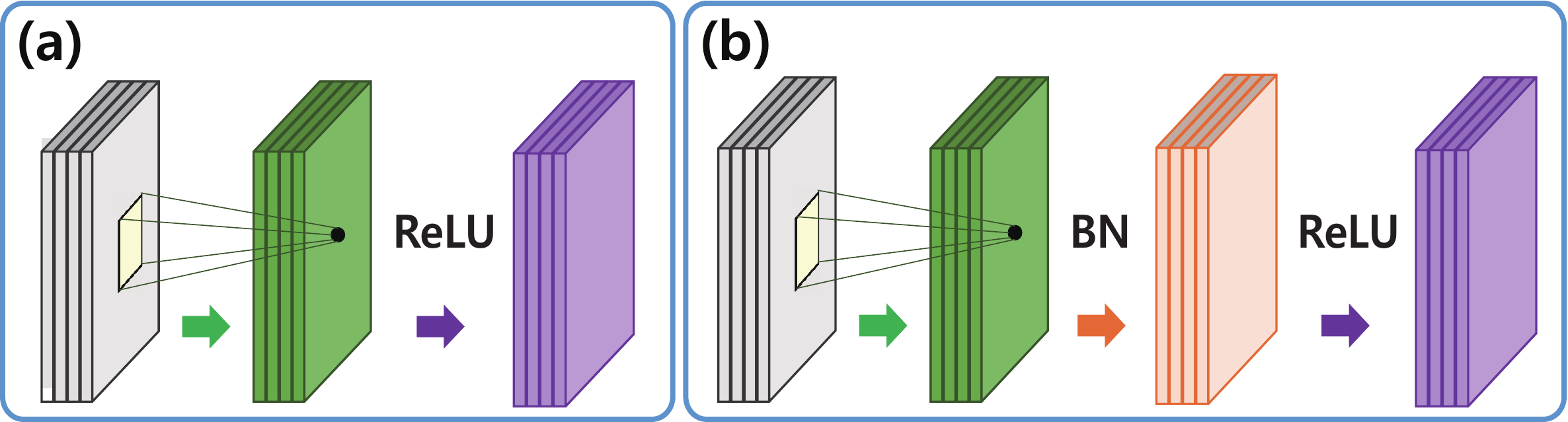}
\caption{Basic unit of a deep CNN: (a) convolution layer and non-linearity (ReLU) layer, and (b) convolution layer, batch normalization (BN) layer and non-linearity (ReLU) layer}
\label{fig:basic_unit_of_CNN}
\end{center}
\end{figure}

For the last few years, various types of deep CNNs  have been developed. 
Several key components of a recent deep CNN include batch normalization \cite{ioffe2015batch}, bypass connection \cite{kim2016accurate}, and a contracting path \cite{ronneberger2015u}. 
These were developed to efficiently train the CNN and improve its performance capabilities. 
For example, the authors of one study \cite{ioffe2015batch} found that the change in the distribution of network activations, called the internal covariate shift, was one of the main culprits of a slow training rate. 
Thus, reducing the internal covariate shift improves the training and minimizes the data overfitting problem as a by-product. 
More specifically, the training data set is subdivided into basic data units designated as minibatches, and batch normalization is then performed as follows,
\begin{equation}
\begin{array}{ll}
	\mu_{\textit{B}} = \dfrac{1}{m} \Sigma^{m}_{i=1} x_i & \quad\quad \mbox{:batch mean} \\
	\sigma^2_{\textit{B}} = \dfrac{1}{m} \Sigma^{m}_{i=1} (x_i - \mu_{\textit{B}})^2 & \quad\quad \mbox{:batch variance} \\
	\hat{x}_i = \dfrac{x_i - \mu_{\textit{B}}}{\sqrt{\sigma^2_{\textit{B}}+\epsilon}} & \quad\quad \mbox{:normalize} \\	
	y_i \leftarrow \gamma \hat{x}_i + \beta \equiv \textit{BN}_{\gamma , \beta}(x_i) & \quad\quad \mbox{:scale and shift}
\end{array}
\end{equation}
where $m$ denotes the batch size, and $\gamma$ and $\beta$ are learnable parameters.
A layer of CNN including batch normalization can then be represented by
\begin{equation}
	y = f(BN(W x)),
\end{equation}
where $BN$ denotes batch normalization.
The structure of the basic unit of the network is described in Fig. \ref{fig:basic_unit_of_CNN}(b).

\begin{figure}
\begin{center}
\includegraphics[width=12cm]{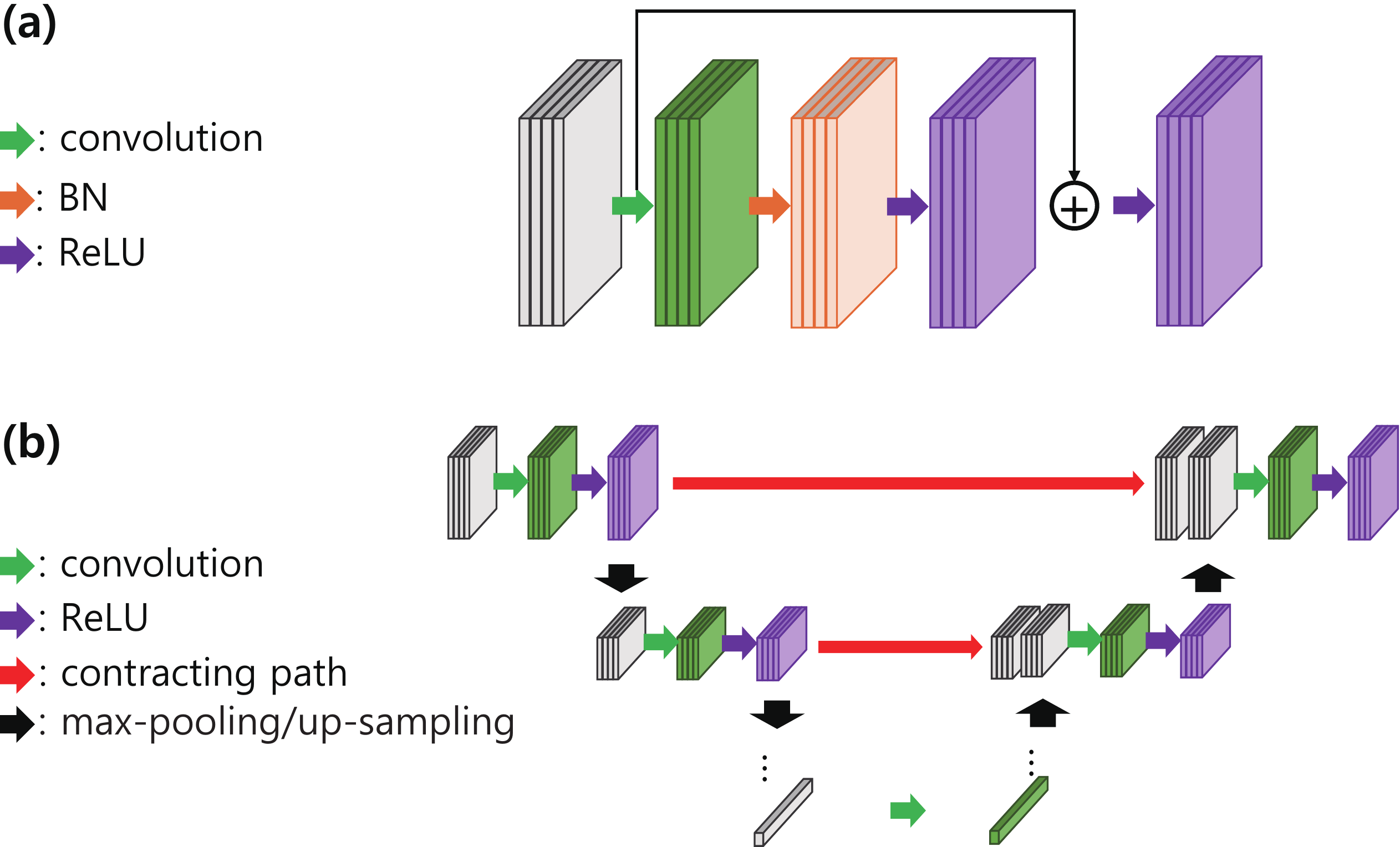}
\caption{(a) Bypass connection as indicated by the black arrow, and (b) contracting path in a U-net as indicated by the red arrow}
\label{fig:skip_and_contract}
\end{center}
\end{figure}

As a network becomes deeper, training of the network becomes more difficult, as the gradients are likely to disappear during an error back-propagation.
To address this issue, the residual learning \cite{he2016deep} was developed, in which a bypass connection \cite{kim2016accurate,mao2016image} and a contracting path \cite{ronneberger2015u} were added to the network.
More specifically, the features processed by bypass connections shown in Fig. \ref{fig:skip_and_contract}(a) carry more image details,  which helps to recover a better image and to secure advantages when back-propagating the gradient.
Similarly, the contract path  described in Fig. \ref{fig:skip_and_contract}(b), originally introduced in U-net \cite{ronneberger2015u} for image segmentation, preserves the details of high-resolution features.
More specifically, a typical CNN has max-pooling (down-sampling) layers such that the information can be lost after passing these layers.
To reduce this phenomenon, high-resolution features from the contracting path are combined with up-sampled output to provide the details of the high-resolution features.

A new CNN architecture was also introduced in which low-frequency image is passed on to the output and learning is performed only for residuals \cite{he2016deep}. 
In terms of training, an adjustable gradient clipping method has been proposed to enable higher learning rates.
It effectively speeds up the convergence for the training procedure.
For the maximum convergence rate, the gradients are truncated to $\left[ \frac{-\theta}{\gamma},\frac{\theta}{\gamma} \right]$, where $\gamma$ is the learning rate.

In the field of medical imaging, CNNs have been used exclusively for medical image analysis and computer-aided diagnosis \cite{suzuki2012pixel}. 
To the best of our knowledge, the proposed method is the first attempt to reduce the noise in low-dose CT images using a CNN. The details of our method are described in Section \ref{sec:method}.

\section{Method}
\label{sec:method}

The proposed network was motivated by the following observations: 
1) a directional wavelet transform such as a contourlet \citep{zhou2005nonsubsampled} can efficiently decompose the directional components of noise to facilitate easier training of a deep network; 
and 2) low-dose CT images have complex noise, and a CNN has great potential to remove such noise;
and 3) a deep neural network is ideal to capture various types of information from a large amount of training data.

\subsection{Contourlet transform}

\begin{figure}
\begin{center}
\includegraphics[width=16cm]{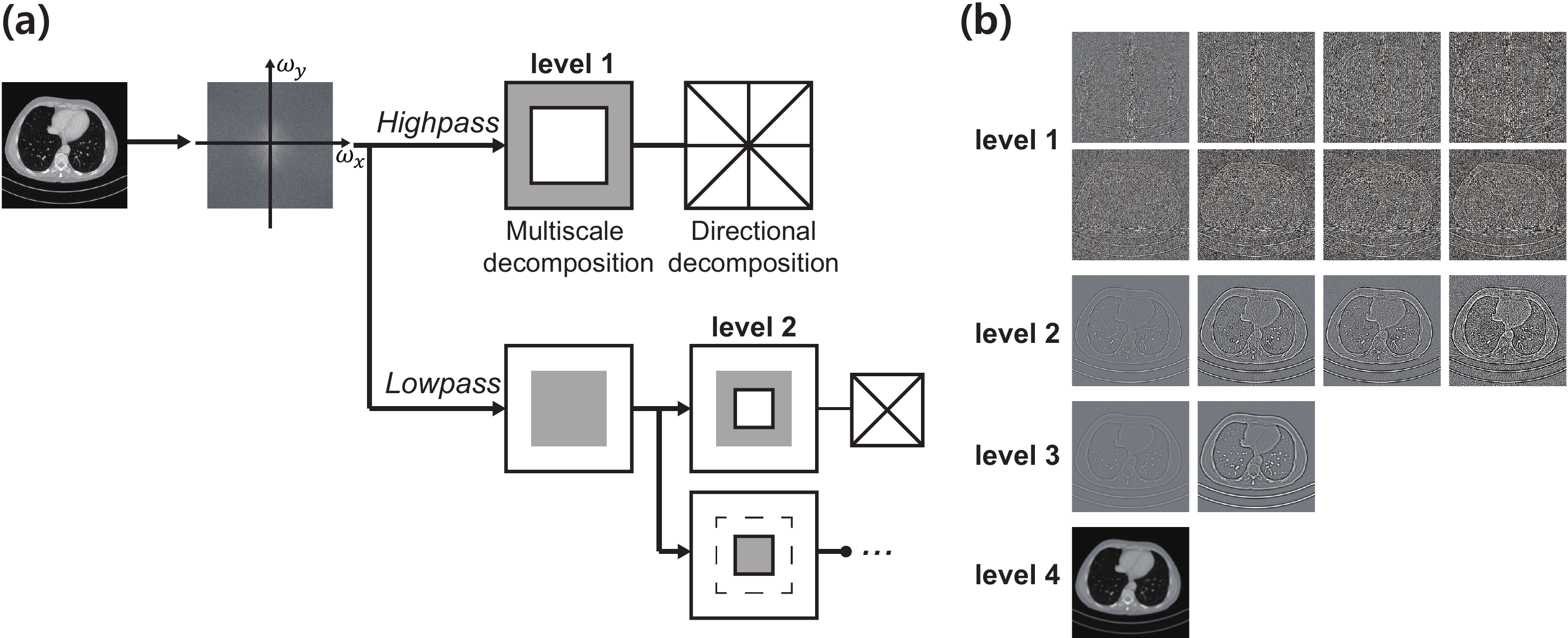}
\caption{Non-subsampled contourlet transform: 
(a) Scheme of  contourlet transform. 
First, the image is split into high-pass and low-pass subbands.
Then, non-subsampled directional filter banks divide the high-pass subband into directional subbands.
This process is repeated in the low-pass subband.
(b) Examples of the non-subsampled contourlet transform of a low-dose CT image.
There are four levels, one each with  eight, four, two, and one directional subbands.}
\label{fig:contourlet}
\end{center}
\end{figure}

The contourlet transform consists of multiscale decomposition and directional decomposition.
The non-subsampled contourlet transform is a shift-invariant version of the contourlet transform which consists of non-subsampled pyramids and non-subsampled directional filter banks, as shown in Fig. \ref{fig:contourlet}(a) \citep{zhou2005nonsubsampled}.
This filter bank does not have down-sampling or up-sampling and is therefore shift-invariant.
Specifically, for a given a high-pass filter $H_1(z)$ and low-pass filter $H_0(z)$, non-subsampled pyramids are constructed by iteration of the filter banks.
More specifically, the $k$th level pyramid is expressed by
\begin{equation}
	H^{eq}_{n}(z) = \left\{ \begin{array}{l l}
		H_1(z^{2^{n-1}})\Pi^{n-2}_{j=0} H_0(z^{2^j}), & \quad 1 \leq n < 2^k \\
		\Pi^{n-1}_{j=0} H_0(z^{2^j}), 				 & \quad n=2^k
	\end{array}
	\right.
\end{equation}
Directional filter banks are then applied to the high-pass subbands to divide them into several directional components.

The scheme of the  contourlet transform and several examples are described in Fig. \ref{fig:contourlet}.
Here, each subband is shown with the same intensity range.
Levels 1, 2, and 3 have high-frequency components of a low-dose CT image, such as edge information and noise.
Note that the streaking noise between the bones is shown in the high-frequency bands.

\subsection{Network architecture}

\begin{figure}[!h]
\begin{center}
\includegraphics[width=16cm]{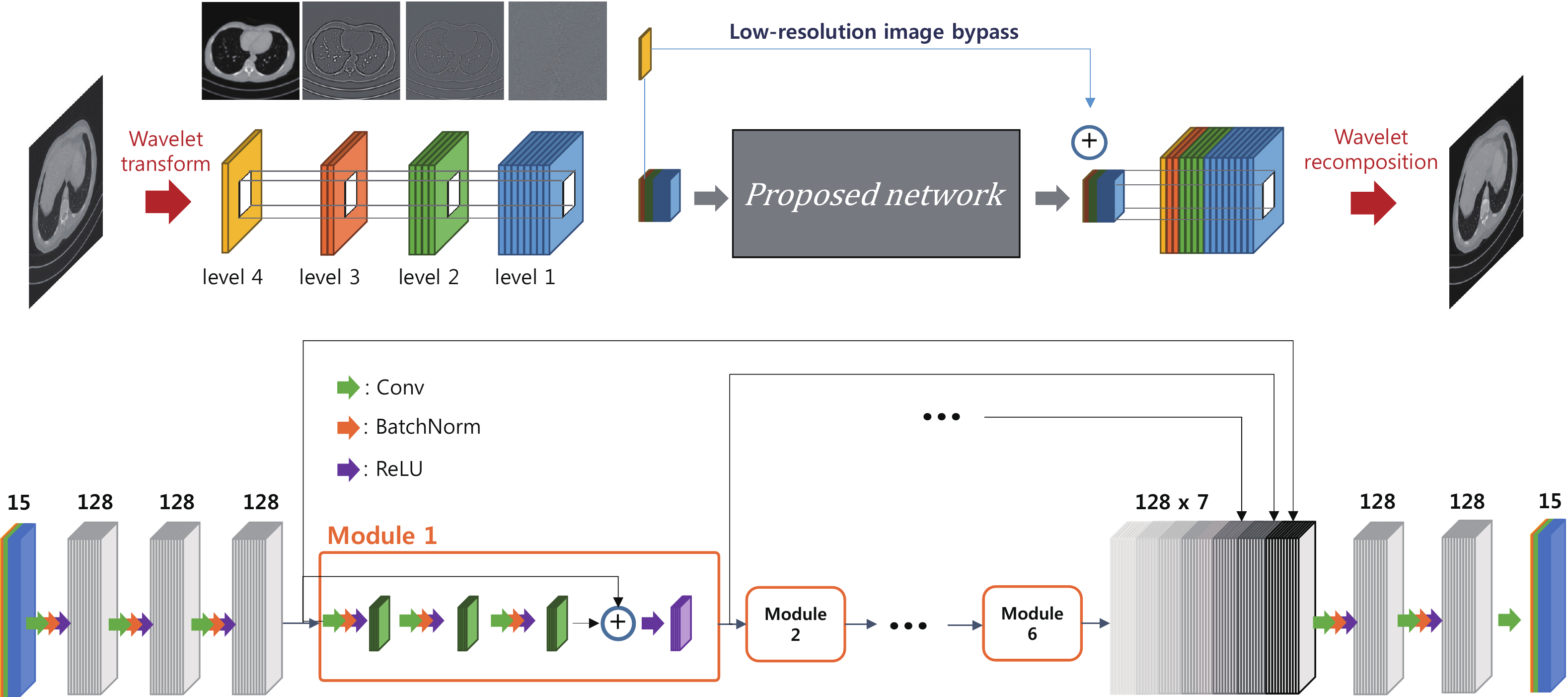}
\caption{Proposed deep convolutional neural network architecture for wavelet domain de-noising}
\label{fig:proposed_network}
\end{center}
\end{figure}

Accordingly, in contrast to the conventional CNN-based denoiser \cite{mao2016image,chen2015learning}, our deep network was designed as a  de-noising approach for wavelet coefficients, as shown in Fig. \ref{fig:proposed_network}.
This idea is closely related to classical de-noising approaches using wavelet shrinkage \cite{donoho1995noising}, but instead of directly applying a closed-form shrinkage operator, the inter- and intra- scale correlations are exploited using a trainable shrinkage operator that transforms noisy wavelet coefficients into clean ones.

More specifically, as shown in Fig. \ref{fig:proposed_network}, an input noisy image is initially decomposed  into four decomposition levels using a contourlet transform \cite{zhou2005nonsubsampled} with a total of  15 channels (8, 4, 2, and 1 for levels 1, 2, 3 and 4, respectively) being generated.
Given that undecimated multi-level contourlet transform is spatially invariant, the noisy wavelet coefficients can be processed in a patch-by-patch manner using a convolution operator.
Here, each patch consists of 55$\times$55 square regions from 15 channels, resulting in the total size of 55$\times$55$\times$15.
Patch-based image denoising techniques commonly use a $50\times 50$ patch, which is capable of describing the noise distribution and containing image information. 
We compared the performance of our network with a patch size between 40 and 60. 
The peak signal-to-noise ratio (PSNR) value of reconstruction images which definition is provided in the Sec. \ref{sec:image_metric}  and the reconstruction time guided the proper patch size, i.e., 55.

In order to make the training more effective, our network takes full advantage of residual learning \cite{he2016deep}.
First, the low-frequency wavelet coefficients were bypassed and later added with the denoised wavelet coefficients, which significantly reduced the unnecessary load on the network.
Because low-frequency images from low-dose image and routine-dose images are nearly equal, learning is not necessary during the training step.
Furthermore, there are other types of internal bypass connections of the network, as shown in Fig. \ref{fig:proposed_network}.
These internal/external bypass connections help overcome the difficulty of training the deep network, resulting in better de-noising performance.

In particular,  the proposed network contains 24 convolution layers, followed by a batch normalization layer and a ReLU layer for each convolution layer except the last one.
Then, 128 sets of 3$\times$3$\times$15 convolution filters are used on the first layer to create 55$\times$55$\times$128 channels, after which 128 sets of 3$\times$3$\times$128 convolution filters are used in the subsequent layers.
Our network consists of six modules, with each module consisting of a bypass connection and three convolution layers. 
In addition, our network has a channel concatenation layer \cite{ronneberger2015u} which stacks the inputs of each module in the channel dimension.
This allows the gradients to be back-propagated over a variety of  paths, enabling  faster end-to-end training.

\subsection{Network training}

Network training was performed by minimizing the loss function \eqref{eq:min} with an additional $l_2$ regularization term for the network parameters.
The regularization parameter $\lambda$ varies in the range of $[10^{-5},10^{-3}]$, and the performance of the proposed network was not sensitive to the choice of $\lambda$.
In fact, our network showed similar performance regardless of $\lambda$ when it varied in the range of $[10^{-5},10^{-3}]$. 

Minimization of the cost function was performed by means of conventional error back-propagation with the mini-batch stochastic gradient descent (SGD)  \cite{rumelhart1986learning,zhang2004solving,cotter2011better} and the gradient clipping method.
The convolution kernel weights were initialized using random Gaussian distributions.
In the SGD, the initial learning rate ($\gamma$) equal to the gradient step size was set to $0.01$, and it continuously decreased  to $10^{-5}$.
The gradient clipping method in the range of $[-10^{-3}, 10^{-3}]$ was used to facilitate the use of a high learning rate in the initial training steps. 
Doing so allows rapid convergence and avoids the gradient explosion problem.
If the learning rate is high, the speed of convergence is fast. 
However, this introduces the gradient explosion problem. 
When we attempted to find an optimal environment setting for a stable learning process, we found that the gradient should be held within the range of $[-10^{-3}, 10^{-3}]$ to prevent gradients from exploding. 

For the mini-batch SGD, the size of the mini-batch was ten, which indicates that ten randomly selected sets of  wavelet coefficients corresponding to 55$\times$55$\times$15 block are used as batches for training.
Furthermore, with regard to data augmentation, the training CT images were randomly flipped, or rotated.
The proposed method was implemented with  MatConvNet \cite{vedaldi2015matconvnet} on MATLAB.
The network training environments are described in Table \ref{table:training_environment}.

\begin{table}[!h]
\centering
\begin{tabular}{ll}
\hline \hline
Training Environment &Specification \\ \hline
Contourlet transform levels and channels &1, 2, 3, 4 levels and 8, 4, 2, 1 channels \\
Patch size &55 $\times$ 55 pixels \\
Number of channels in the network & 128 channels \\
Convolution layer filter size in X-Y domain & 3 $\times$ 3 \\
Learning rate range & $[10^{-5}, 10^{-2}]$ \\
Gradient clipping range & $[-10^{-3}, 10^{3}]$ \\
Size of mini-batch & 10 \\
\hline \hline
\end{tabular}
\caption{Hyper-parameters in the proposed network}
\label{table:training_environment}
\end{table}

\begin{table}[!h]
\centering
\begin{tabular}{c|c|c|c|c|c|c}
\hline \hline
\multirow{2}{*}{Patient ID}	&\multirow{2}{*}{Number of slices} &Size of FOV &\multirow{2}{*}{KVP} 
&Exposure time &\multicolumn{2}{c}{X-ray tube current [mA]} \\ \cline{6-7}
 		& 		&[mm]	& 		&[ms]	&Routine dose 	&Quarter dose 	\\ \hline \hline
L067 	&$310$ 	&$370$ 	&$100$ 	&$500$ 	&$234.1$ 		&59.2 			\\ \hline
L097 	&$500$ 	&$430$ 	&$120$ 	&$500$ 	&$327.6$  		&82.9			\\ \hline
L109 	&$254$ 	&$400$ 	&$100$ 	&$500$ 	&$322.3$ 		&79.2 			\\ \hline
L143 	&$418$ 	&$440$ 	&$120$ 	&$500$ 	&$416.9$  		&105.5			\\ \hline
L192 	&$370$ 	&$380$ 	&$100$ 	&$500$ 	&$431.6$ 		&109.2			\\ \hline
L286 	&$300$ 	&$380$ 	&$120$ 	&$500$ 	&$328.9$  		&82.2			\\ \hline
L291 	&$450$ 	&$380$ 	&$120$ 	&$500$ 	&$322.7$ 		&81.7 			\\ \hline
L310 	&$340$ 	&$380$ 	&$120$ 	&$500$ 	&$300.0$  		&73.7			\\ \hline
L333 	&$400$ 	&$400$ 	&$100$ 	&$500$ 	&$348.7$  		&88.2			\\ \hline
L506  	&$300$ 	&$380$ 	&$100$ 	&$500$ 	&$277.7$  		&70.2			\\ 
\hline \hline
\end{tabular}
\caption{Training data set specifications:
Size of the FOV in units of [mm], exposure time in units of [ms], and X-ray tube current in units of [mA]}
\label{table:training_data set}
\end{table}

\begin{table}[!h]
\centering
\begin{tabular}{c|c|c|c|c|c}
\hline \hline
\multirow{2}{*}{Patient ID}	&Number of slices &Size of FOV &\multirow{2}{*}{KVP} 
&Exposure time &X-ray tube current [mA] \\ 
 		&(3mm slice thickness) 		&[mm]	& 		&[ms]	&(Quarter dose)	\\ \hline \hline
L008 	&$110$ 	&$360$ 	&$100$ 	&$500$ 	&68.8 			\\ \hline
L031 	&$111$ 	&$460$ 	&$120$ 	&$500$ 	&70.7			\\ \hline
L057 	&$118$ 	&$380$ 	&$100$ 	&$500$ 	&78.5			\\ \hline
L061 	&$83$ 	&$380$ 	&$100$ 	&$500$ 	&139.4			\\ \hline
L072 	&$119$ 	&$340$ 	&$100$ 	&$500$ 	&74.1			\\ \hline
L106 	&$98$ 	&$380$ 	&$100$ 	&$500$ 	&86.4			\\ \hline
L123 	&$127$ 	&$500$ 	&$120$ 	&$500$ 	&135.8 			\\ \hline
L136 	&$110$ 	&$400$ 	&$120$ 	&$500$ 	&74.9			\\ \hline
L205 	&$80$ 	&$320$ 	&$100$ 	&$500$ 	&73.4			\\ \hline
L243 	&$116$ 	&$380$ 	&$100$ 	&$500$ 	&96.4 			\\ \hline
L254 	&$107$ 	&$380$ 	&$100$ 	&$500$ 	&65.1			\\ \hline
L433 	&$97$ 	&$400$ 	&$120$ 	&$500$ 	&114.7 			\\ \hline
L541 	&$107$ 	&$350$ 	&$100$ 	&$500$ 	&71.7			\\ \hline
L548 	&$95$ 	&$340$ 	&$100$ 	&$500$ 	&66.3			\\ \hline
L554 	&$122$ 	&$400$ 	&$120$ 	&$500$ 	&93.8			\\ \hline
L562 	&$98$ 	&$400$ 	&$100$ 	&$500$ 	&121.5 			\\ \hline
L581 	&$91$ 	&$380$ 	&$100$ 	&$500$ 	&115.7			\\ \hline
L593 	&$107$ 	&$360$ 	&$100$ 	&$500$ 	&46.2			\\ \hline
L631 	&$104$ 	&$320$ 	&$100$ 	&$500$ 	&55.8			\\ \hline
L632  	&$101$ 	&$380$ 	&$120$ 	&$500$ 	&85.9			\\ 
\hline \hline
\end{tabular}
\caption{Test data set specifications:
Size of the FOV in units of [mm], exposure time in units of [ms], and X-ray tube current in units of [mA]}
\label{table:test_data set}
\end{table}

\subsection{Data Set}

In the 2016 CT low-dose Grand Challenge, only abdominal CT images were provided as the training data set, as shown in Table~\ref{table:training_data set}.
The training data sets consist of normal-dose and quarter-dose CT fanbeam reconstruction data from ten patients.
The data is composed of 3-D CT projection data from 2304 views and the total number of slices was 3642.
In the Challenge, the test data, consisting only of quarter-dose exposure images, were also provided, the specifications of which are shown in Table~\ref{table:test_data set}.
This data consists of 2101 slices from 20 patients.
In the Challenge, radiologists evaluated only these results; hence, we also provide some of the test data reconstruction results in this paper.

For training and test data, we generated the image data from projection data.
Specifically, with the given raw projection data acquired by a 2D cylindrical detector and a helical conebeam trajectory using a z-flying focal spot \cite{flohr2005image}, they were approximately converted into conventional fanbeam projection data using a single-slice rebinning technique \cite{noo1999single}.
In particular, we considered the movement of the X-ray source of the z-flying focal spot in the rebinning technique, and the final rebinned data were generated with a slice thickness of 1mm.
From the rebinned data, $512\times 512$ CT images were reconstructed  using a conventional filtered backprojection algorithm.

The CT images reconstructed from the normal-dose data were then used as the ground truth images, the quarter-dose images were used as noisy input, and the  mapping between them was learned.
In each epoch, all  training data sets are used once to update the weights.
However, the computer memory was not sufficient to use the entire training data set.
Therefore, we randomly extracted 200 slices from the 3642 slices of training data set and changed them in an interval of 50 epochs.
Here, epoch refers to how often the weights are updated with 200 slices.

For the first submission of the 2016 Low dose CT Grand Challenge, our network was trained with 1-mm-thick CT images from the data of ten patients.
The final submitted images were formed with a thickness of 3mm by averaging three adjacent images with thicknesses of 1mm.
Considering that 3mm thickness results were used for the evaluation in the Low-Dose CT Grand Challenge, here we also re-trained the proposed network with a different strategy. 
In particular, we initially obtained the 3mm average images of adjacent three 1mm slice images to construct the training data.
The proposed network was then re-trained with the 3mm CT images using the average CT data at a thickness of 3mm.

\subsection{Image Metrics}
\label{sec:image_metric}

For a quantitative assessment, we used a data set from a patient in Table \ref{table:training_data set} and calculated the image metrics, specifically the peak signal-to-noise ratio (PSNR) and the normalized root mean square error (NRMSE) values.
These metrics are defined in terms of the mean square error (MSE), which is defined as
\begin{equation}
	MSE = \dfrac{1}{mn} \sum_{i=0}^{m-1} \sum_{j=0}^{n-1} [ \Yb(i,j) - \Xb(i,j)]^2,
\end{equation}
where $\Yb$ is a normal-dose (ground truth) image and $\Xb$ is the reconstruction from the noisy input.
The defined value of the PSNR is expressed by
\begin{eqnarray}
	PSNR &=& 10 \cdot \log_{10} \left(\dfrac{MAX_{\Yb}^2}{MSE}\right) \\
		 &=& 20 \cdot \log_{10} \left(\dfrac{MAX_{\Yb}}{\sqrt{MSE}}\right), 
\end{eqnarray}
where $MAX_{\Yb}$ is the maximum value of image $\Yb$, and the NRMSE is defined using the square root of the mean square error (RMSE),
\begin{eqnarray}
	RMSE &=& \sqrt{MSE}, \\
	NRMSE &=& \dfrac{RMSE}{MAX_{\Yb} - MIN_{\Yb}},
\end{eqnarray}
where $MIN_{\Yb}$ is the minimum value of image $\Yb$.
The data consisting of normal-dose images was used as the ground truth and the denoised images from quarter-dose images were compared to the calculated above-mentioned metrics.

\begin{figure}[!h]
\begin{center}
\includegraphics[width=11cm]{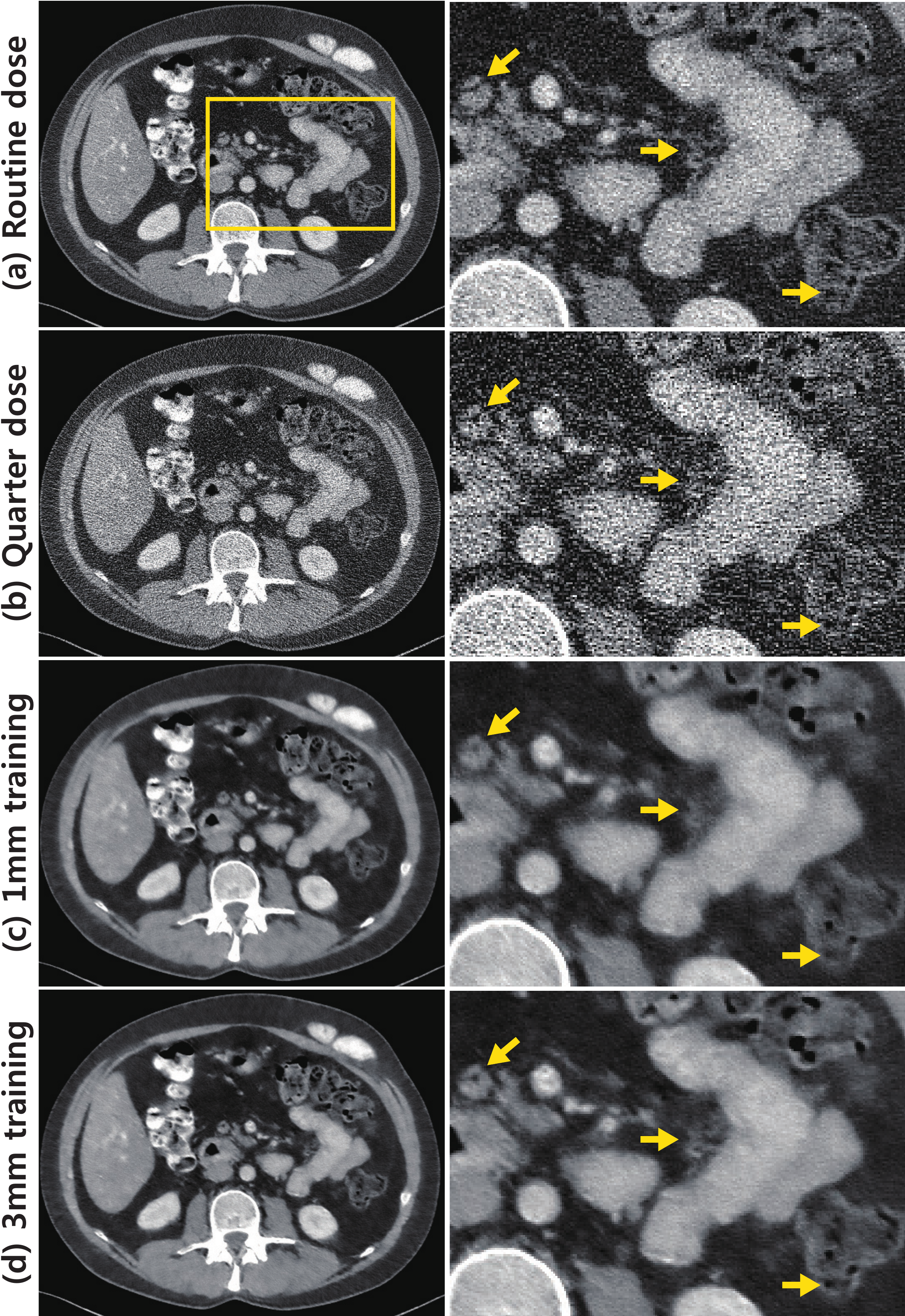}
\caption{Reconstruction results from the training data `L291':
(a) routine-dose image, (b) quarter-dose image, and the results with (c) the proposed network trained with 1mm slices followed by 3mm averaging, and (d) the proposed network trained with 3mm slices. The second column shows enlarged images from the yellow boxes.
Yellow arrows denote the image details.
The intensity range was set to (-160,240) [HU].}
\label{fig:result_training_3mm}
\end{center}
\end{figure}

\begin{figure}[!h]
\begin{center}
\includegraphics[width=11cm]{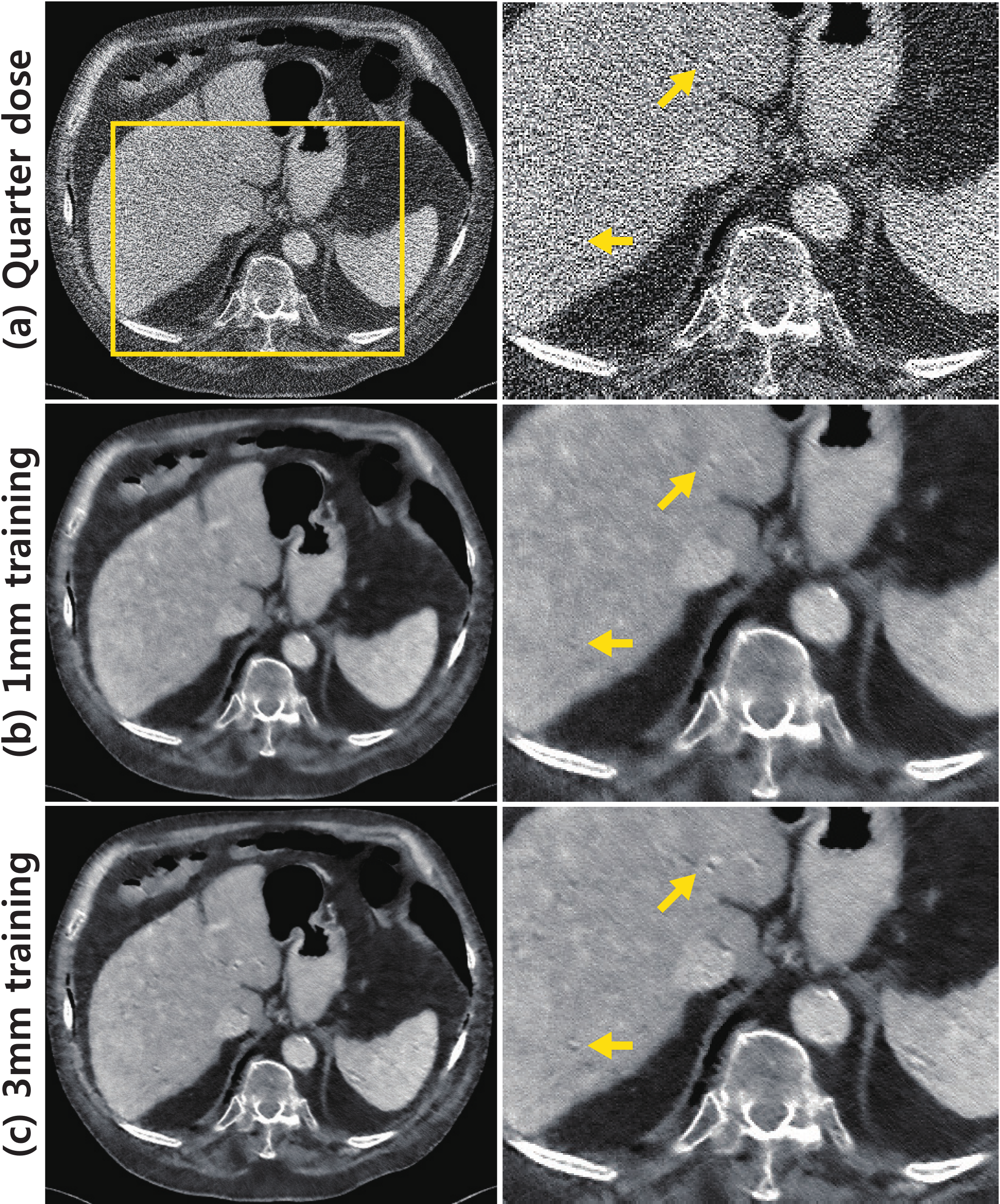}
\caption{Reconstruction results from the test data `L031':
(a) quarter-dose image, and the results by (b) the proposed network trained with 1mm slices followed by 3mm averaging, and
(c) the proposed network trained with 3mm slices. The second column shows enlarged images from the yellow boxes.
Yellow arrows indicate the image details.
The intensity range was set to (-160,240) [HU].}
\label{fig:result_test_3mm}
\end{center}
\end{figure}

\section{Experimental results}
\label{sec:result}

\subsection{Slice thickness}

First, the comparative results from the 1mm and 3mm training data are shown in Fig. \ref{fig:result_training_3mm} and Fig. \ref{fig:result_test_3mm}.
The denoised images through the newly trained network with 3mm images preserve the fine image details better than those of the previous network.
In particular, the image edges, such as the boundaries and details of the organs, become clearer.
Although the denoised images, trained with 1mm slices, retained the details of the regions with lesions and significantly suppress the streaking artifacts, we found that the denoised images appeared somewhat blurred and that some high-frequency textures were often lost.
This limitation resulted from the fact that the normal-dose CT images with a thickness of 1mm also contain noise, which reduces the accuracy of the supervised learning process.

With regard to  the  computation complexity, the proposed deep CNN framework is very advantageous;
we showed that the average processing time for a 512 x 512 pixel CT image is approximately  {1.6 seconds per slice} for the MATLAB implementation with a dual-graphical processing unit (NVidia GeForce Titan 6GB).
Accordingly, even a whole-body CT scan data can be processed by our method in a time frame of 2.1 $\sim$ 3.3 minutes, even when implemented on MATLAB.
The calculation can be optimized further by optimizing the network architecture and the parallel computations and using a dedicated software platform for deep learning instead of MatConvNet.

\begin{figure}[!h]
\begin{center}
\includegraphics[width=16cm]{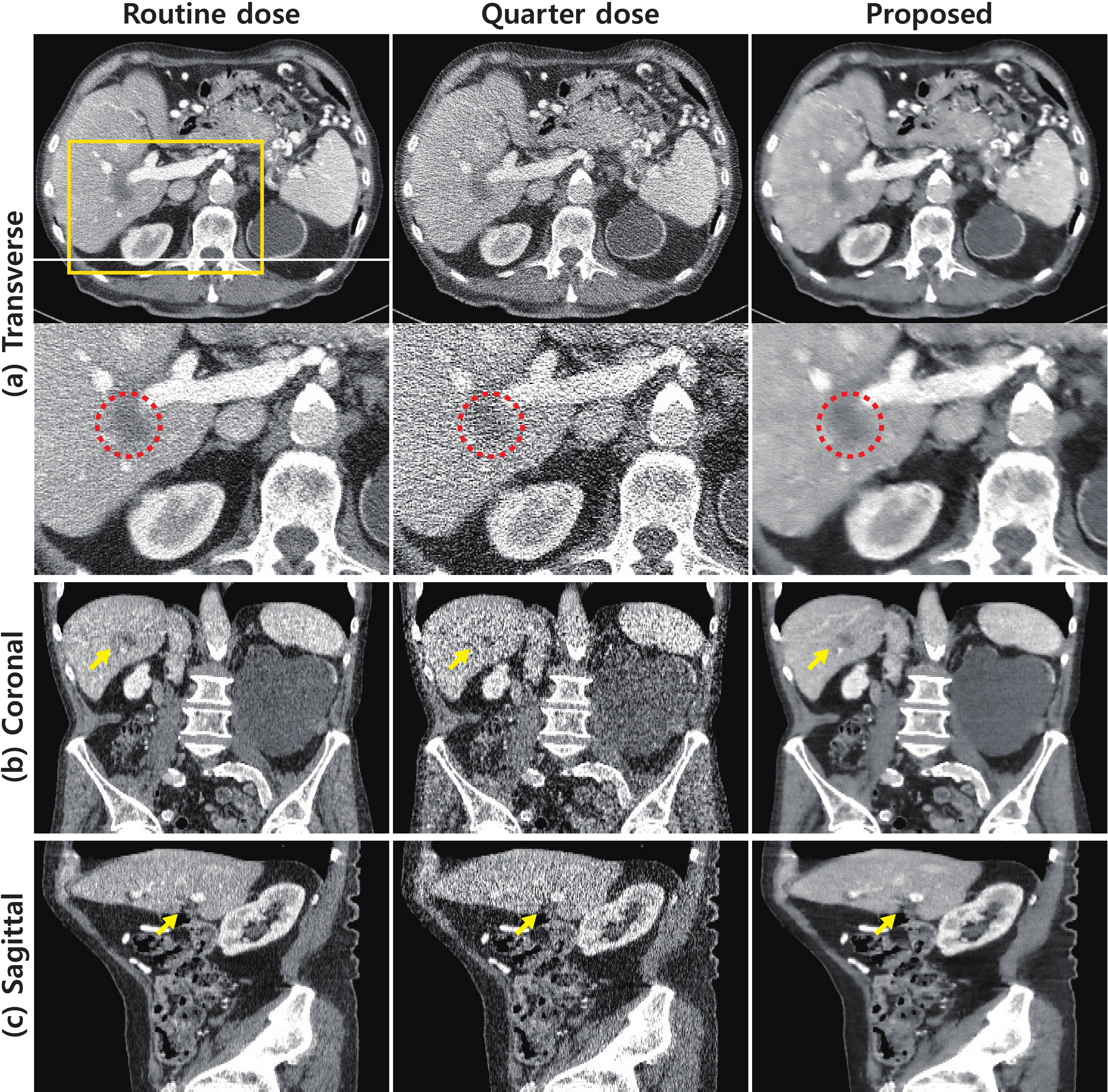}
\caption{X-ray CT images from training data set `L506'.
Routine-dose images are in the first column, quarter-dose images are in the second column, 
and denoised images using the proposed algorithm are in the third column.
(a) Transverse images. Enlarged images within the yellow box in the second row.
The lesion is marked by red dashed circles.
(b) Coronal images and (c) Sagittal images.
The intensity range was set to (-160,240) [HU] (Hounsfield unit).}
\label{fig:result_training}
\end{center}
\end{figure}

\begin{figure}[!h]
\begin{center}
\includegraphics[width=11cm]{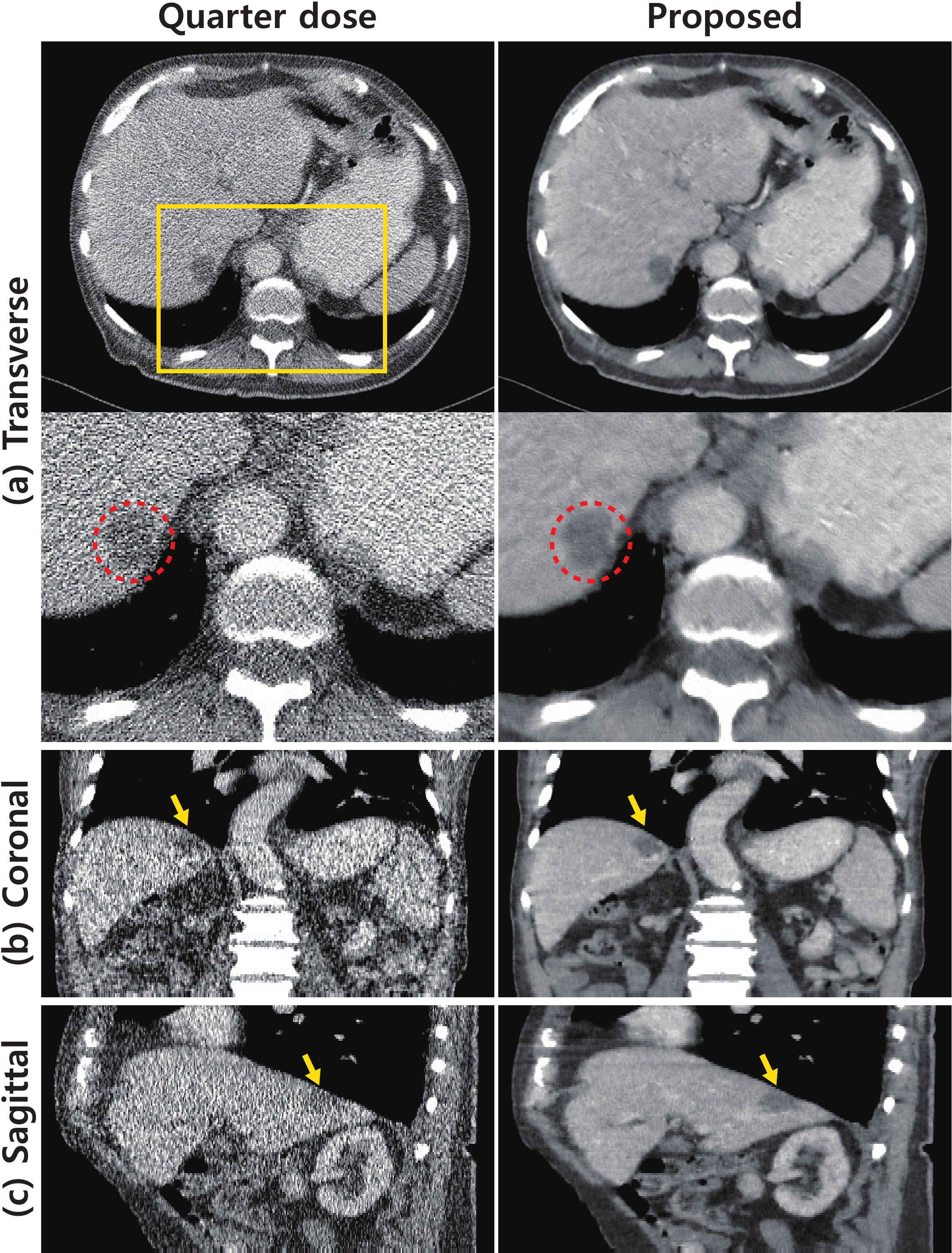}
\caption{X-ray CT images from test data set `L057'.
Quarter-dose images are in the first column and denoised images using the proposed algorithm are in the second column.
(a) Transverse images. Enlarged images within the yellow box in the second row.
The lesion is marked by red dashed circles.
(b) Coronal images and (c) Sagittal images. Yellow arrow indicate the lesion.
The intensity range was set to (-160,240) [HU] (Hounsfield unit).}
\label{fig:result_test}
\end{center}
\end{figure}

\begin{figure}[!h]
\begin{center}
\includegraphics[width=11cm]{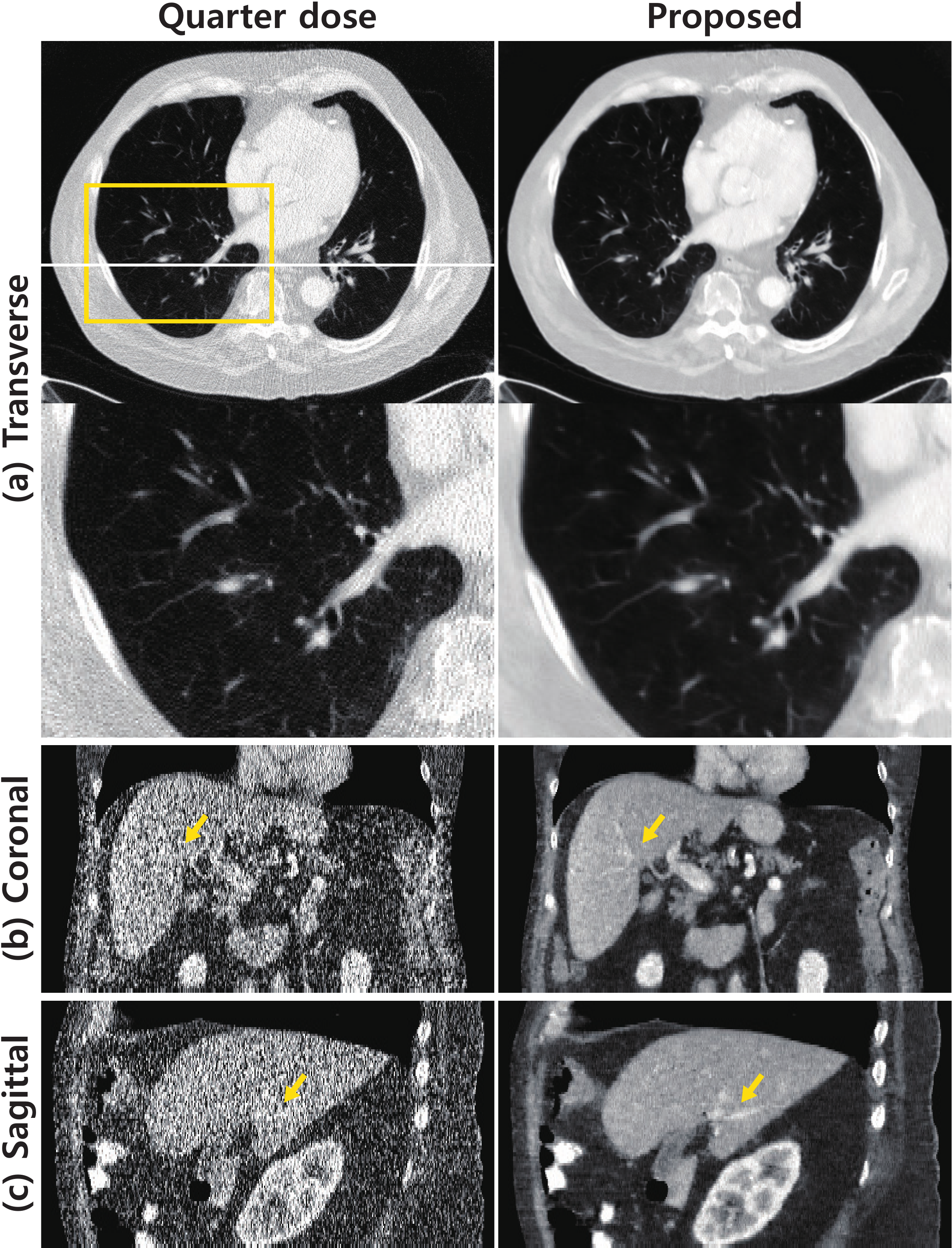}
\caption{X-ray CT images from test data set `L031'.
Quarter-dose images are in the first column and denoised images using the proposed algorithm are in the second column.
(a) Transverse images. Enlarged images within the yellow box in the second row.
The intensity range has been adjusted to highlight the details of the lung.
(b) Coronal images and (c) Sagittal images. 
Yellow arrows and circles indicate the details of the liver.}
\label{fig:lung_and_liver}
\end{center}
\end{figure}

For a subjective evaluation, 
Fig. \ref{fig:result_training} shows the denoised images from the data set of one patient with a normal-dose and a quarter-dose.
Organs such as the liver can be seen in these images.
The results of the proposed network preserved the textures of the liver such that a better determination of the location of the lesion was possible, as highlighted by the red dashed circle.
The coronal and sagittal presentation of the results is also shown in Fig. \ref{fig:result_training}.
Yellow arrows indicate the regions with high noise levels.
The proposed network can remove a wide range of noise levels while maintaining the edge information.

Fig. \ref{fig:result_test} shows the denoised images of data from one patient from among the 20 patients with only quarter-dose data.
The trained network is applied to the test data and its denoising performance is demonstrated by the determination of the location of the lesion, as indicated by the red dashed circle.
The coronal and sagittal presentation of the results is also shown in Fig. \ref{fig:result_test}.
Yellow arrows indicate the location of the lesion, providing a better view to assist with the understanding of the condition of the patient.

To demonstrate the detail preservation performance capabilities of the proposed network, we examine the results of another test data set, as shown in Fig. \ref{fig:lung_and_liver}.
The lung and other organs are visible in these images.
The proposed network was able to describe the details of the lung structure.
We also observed that the vessels in the liver are clearly reconstructed, as indicated by the yellow arrows in the coronal and sagittal presentation of the results.

\begin{figure}[!h]
\begin{center}
\includegraphics[width=16cm]{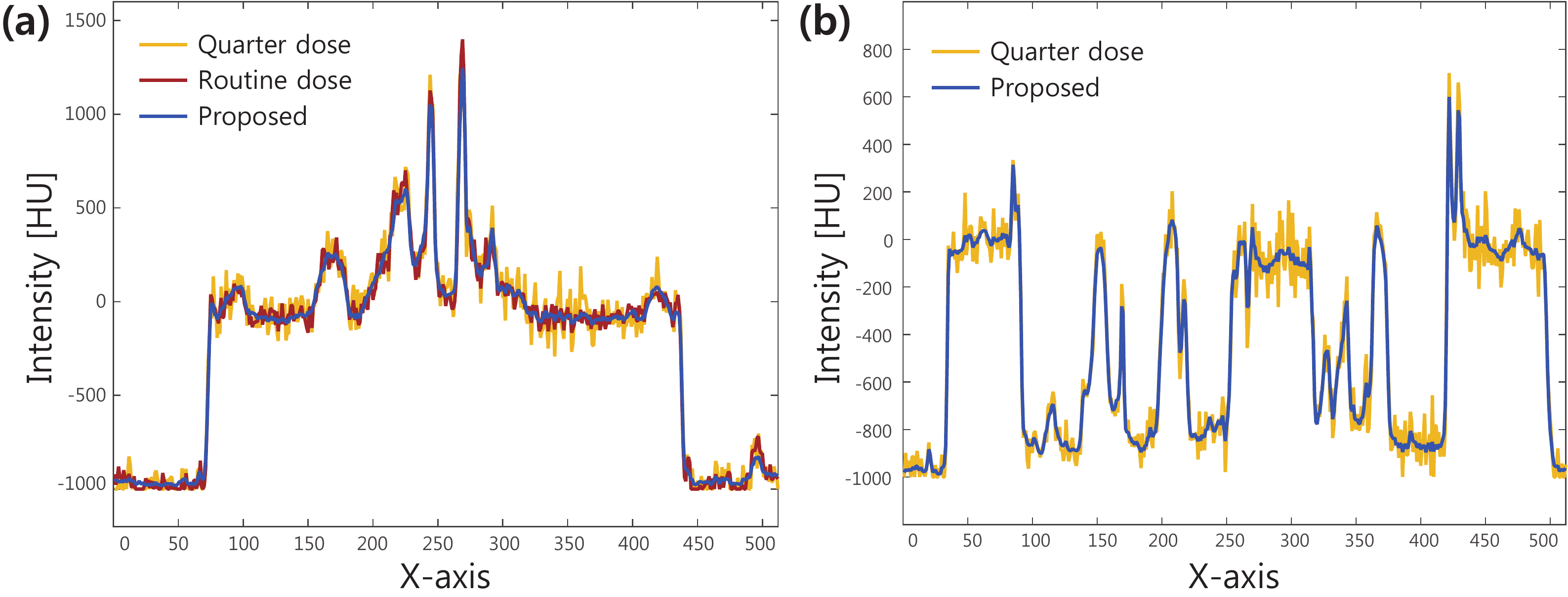}
\caption{(a) Intensity profile along the line in Fig. \ref{fig:result_training}, and
(b) Intensity profile along the line in Fig. \ref{fig:lung_and_liver}}
\label{fig:result_profile}
\end{center}
\end{figure}

Profiles of the results are shown in Fig. \ref{fig:result_profile} from both the training and the test data sets. 
Here, the corresponding positions of the profiles in Figs. \ref{fig:result_training} and \ref{fig:lung_and_liver} are indicated by white solid lines.
In  Fig. \ref{fig:result_profile}(a), the proposed network suitably reduces noise and describes the peak points.
Moreover, the profiles of the results from the test data set in Fig. \ref{fig:result_profile}(b) show that the proposed network result also feasibly reduces noise in the test data.

\subsection{Role of the wavelet transform}

\begin{figure}[!h]
\includegraphics[width=16cm]{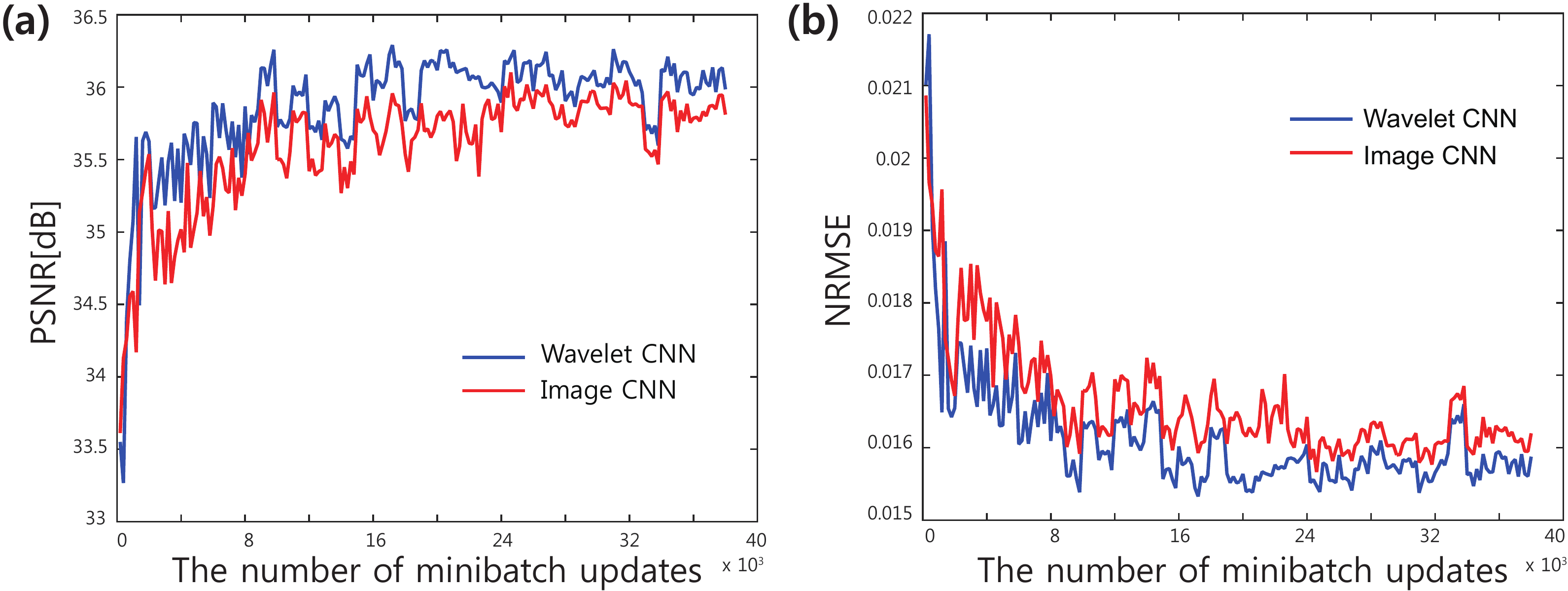}
\caption{Comparison of the proposed network versus the image-domain CNN:
(a) PSNR, and (b) NRMSE}
\label{fig:Wavelet_analysis}
\end{figure}

To verify the role of the wavelet transform, the proposed network was compared with a baseline CNN - an image-based neural network that is identical to the proposed network, except that its input and output layer are images.
The  baseline network was applied to the image patches instead of the local wavelet coefficients.
These two networks were trained with the data from nine patients, and the remaining data from one patient was used for the evaluation. 
During the training process, the degrees of convergence were evaluated with the data from the one other patient. 
Fig. \ref{fig:Wavelet_analysis} shows that our wavelet-based CNN outperforms the image-domain CNN  with regard to the peak signal-to-noise ratio (PSNR) and the normalized root mean square error (NRMSE).

\begin{figure}[!h]
\includegraphics[width=16cm]{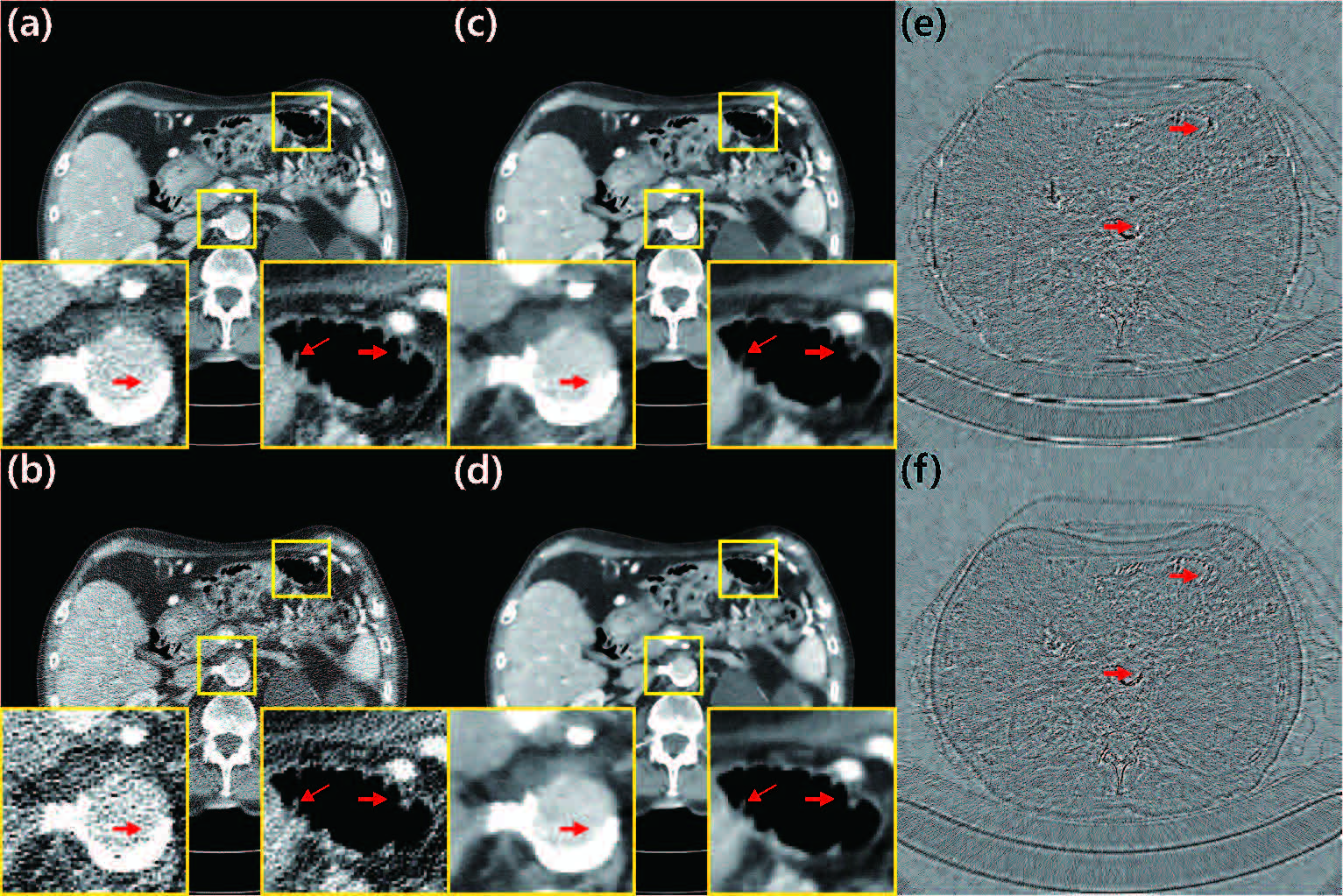}
\caption{Result images of the proposed network and the image-domain CNN:
(a) Routine-dose image, (b) quarter-dose image, (c) image-domain CNN result,
(d) the result from the proposed network, (e) the difference between the routine-dose and the image-domain CNN result, and (f) the difference between routine-dose and the proposed network result.
The  intensity range of the results was set to (-160,240) [HU],
and the  intensity range of the difference image  was set to (-100,100) [HU].}
\label{fig:Wavelet_analysis_result}
\end{figure}

Fig. \ref{fig:Wavelet_analysis_result} illustrates the results of the proposed network and the image-domain CNN.
The performance capabilities of the image-domain CNN were also impressive.
On the other hand, the differences between the two were clearly visible in the images of difference, as they were mainly from the image edges. 
Moreover, many structures  are not recovered in the image-domain CNN, as indicated by the red arrows in the yellow magnified boxes.

\subsection{Analysis of residual learning techniques}

\begin{figure}[!h]
\begin{center}
\includegraphics[width=16cm]{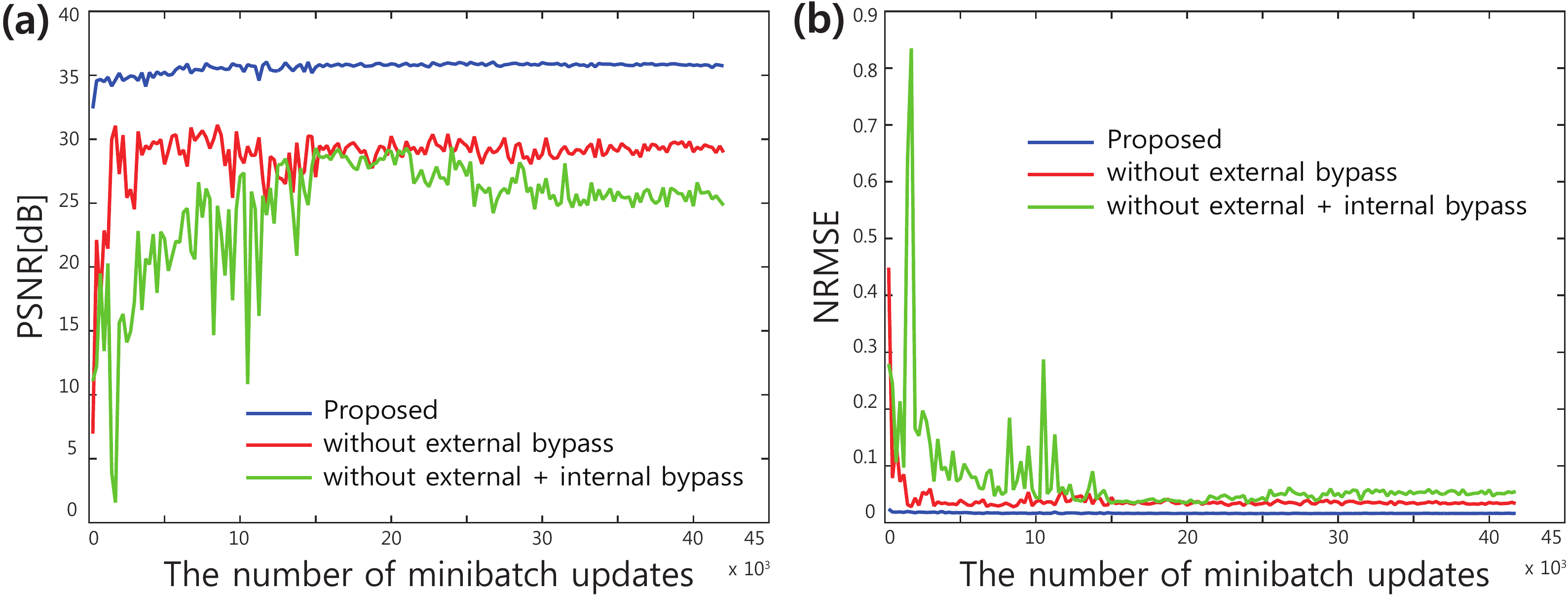}
\caption{Importance of residual learning:
The proposed network is compared to a baseline network structure without an external low-frequency band bypass path and another baseline network without external bypass and internal bypass connections.
Quantification by (a) PSNR, and (b) NRMSE.}
\label{fig:residual_analysis}
\end{center}
\end{figure}

In order to  verify the effect of the residual learning technique,
the proposed network was compared with another baseline CNN with an identical architecture but without residual learning.
The mapping between the contourlet coefficients of the lowest frequency band (level 4) was also learned in the baseline CNN (the red curve in Fig. \ref{fig:residual_analysis}).
The comparison clearly shows the importance of residual learning.

Recall that in the proposed network, the input is added to the output before the ReLU layer in each module, and we combine the output of each module in the last step to form the concatenated layer, as shown in Fig.~\ref{fig:proposed_network}.
Accordingly, we also analyzed the effectiveness of these internal bypass connections used in each module during the last step of the network.
In Fig. \ref{fig:residual_analysis}, the proposed network is also compared to the baseline CNN, as designated by the green line, which does not have an external bypass path for low-frequency coefficients and internal bypass connections.
Fig. \ref{fig:residual_analysis} confirms that our unique residual learning architecture is helpful to train the proposed network.

\section{Discussion}
\label{sec:discussion}

An important advantage of the proposed network over MBIR approaches is that the proposed network can fully utilize a large training data set, if such a data set is available, given that a large number of neural network parameters can be trained more accurately  with more data.
On the other hand, MBIR approaches usually train a single regularization parameter despite the availability of a large training data set.
Therefore, we believe that our method offers a significant advantage over MBIR approaches due to its ability to learn organ, protocol,  and hardware-dependent noise from a larger available training data set.

In the 2016 CT low-dose Grand Challenge, only abdominal CT images with a quarter-dose were provided.
Our network is adapted to a quarter-dose, while the noise distribution changes according to the dosage level.
When we applied it to reconstruction from lower level noise, the denoised images contained blurring artifacts.
Therefore, to enable a low-dose CT at another dose level, additional training with data with a different noise level will be required.
However, transfer learning \cite{taylor2009transfer,yosinski2014transferable,pan2010survey} or domain adaptation\cite{ganin2016domain} can now be considered feasible for use with deep learning.
Thus, a pre-trained network can serve as the start point of training with another data set with different noise levels.
This will reduce the training time and produce better results than training using a new data set only.
Moreover, with regard to natural image de-noising, CNN networks are currently actively investigated for different noise levels.
A previous study \cite{burger2012image} demonstrated the possibility that a neural network can strongly remove various levels of noise.
Similarly, if we have a multiple-dosage data set, we can train the network for different dose levels using all of the training samples from different noise levels.
This type of heterogeneous training was shown to be effective in our recent work on deep-learning-based sparse view CT reconstruction
\cite{han2016deep}.

Projection domain de-noising has also been widely investigated for low-dose CT.
Therefore, we can apply the proposed network to projection data.
However, projection data is related across all angles, making the use of patch processing techniques more difficult.
This difficulty can be overcome by a network with a large receptive field, similar to that in our recent work\cite{han2016deep}, though this goes beyond the scope of the present work.

Finally, some of the radiologists who evaluated our results in the Low-dose CT Grand Challenge mentioned that the texture of our deep-learning reconstruction differs from those of MBIR methods.
Because the textures of images are also important diagnostic features, a new deep-learning method is needed to deal with these problems as a follow-up study to this work.

\section{Conclusion}
\label{sec:conclusion}

In this paper, we introduced a deep CNN framework designed for low-dose CT reconstruction.
It combines a deep convolution neural network with a directional wavelet approach.
We demonstrated that the proposed method has greater de-noising power for low-dose CT and that its reconstruction time is much faster than those of MBIR methods.
The effectiveness of the proposed network was confirmed in the 2016 AAPM Low-Dose CT Grand Challenge.
We believe that the method presented here suggests a new innovative  framework for low-dose CT research.

\section*{Acknowledgement}
The authors would like to thanks Dr. Cynthia McCollough, the Mayo Clinic, the American Association of Physicists in Medicine (AAPM), and grant EB017095 and EB017185 from the National Institute of Biomedical Imaging and Bioengineering for providing the Low-Dose CT Grand Challenge data set.

This work is supported by Korea Science and Engineering Foundation under grant number NRF-2016R1A2B3008104, 
Industrial Strategic technology development program (10072064, Development of Novel Artificial Intelligence Technologies To Assist Imaging Diagnosis of Pulmonary, Hepatic, and Cardiac Diseases and Their Integration into Commercial Clinical PACS Platforms) funded by the Ministry of Trade Industry and Energy (MI, Korea) and
Institute for Information \& Communications Technology Promotion(IITP) grant funded by the Korea government(MSIP) (R0124-16-0002, Emotional Intelligence Technology to Infer Human Emotion and Carry on Dialogue Accordingly).\\

%\bibliography{wavelet_cnn_reference}

%merlin.mbs aapmrev4-1.bst 2010-07-25 4.21a (PWD, AO, DPC) hacked

\end{document}